\documentclass[letterpaper]{article} 
\usepackage{aaai25}  
\usepackage{times}  
\usepackage{helvet}  
\usepackage{courier}  
\usepackage[hyphens]{url}  
\usepackage{enumitem}
\usepackage{amsmath}
\usepackage{booktabs}
\usepackage{graphicx} 
\usepackage{natbib}  
\usepackage{caption} 
\usepackage{algorithm}
\usepackage{algorithmic}

\usepackage{newfloat}
\usepackage{listings}
\DeclareCaptionStyle{ruled}{labelfont=normalfont,labelsep=colon,strut=off} 
\floatstyle{ruled}
\newfloat{listing}{tb}{lst}{}
\floatname{listing}{Listing}
\title{Comparative Analysis of Demonstration Selection Algorithms for \\ LLM In-Context Learning}
\author {
    Dong Shu\textsuperscript{\rm 1},
    Mengnan Du\textsuperscript{\rm 2}
}
\affiliations {
    \textsuperscript{\rm 1}Northwestern University\\
    \textsuperscript{\rm 2}New Jersey Institute of Technology\\
    dongshu2024@u.northwestern.edu, mengnan.du@njit.edu
}

\usepackage{bibentry}

\begin{document}

\maketitle

\begin{abstract}
In-context learning can help Large Language Models (LLMs) to adapt new tasks without additional training. However, this performance heavily depends on the quality of the demonstrations, driving research into effective demonstration selection algorithms to optimize this process. These algorithms assist users in selecting the best $k$ input-label pairs (demonstration examples) based on a given test input, enabling LLMs to in-context learn the relationship between the provided examples and the test inputs. Despite all the proposed demonstration selection algorithms, their efficiency and effectiveness remain unclear. This lack of clarity make it difficult to apply these algorithms in real-world scenarios and poses challenges for future research aimed at developing improved methods. This paper revisits six proposed algorithms, evaluating them on five datasets from both efficiency and effectiveness perspectives. Our experiments reveal significant variations in algorithm performance across different tasks, with some methods struggling to outperform random selection in certain scenarios. We also find that increasing the number of demonstrations does not always lead to better performance, and that there are often trade-offs between accuracy and computational efficiency. Our code is available at \url{https://github.com/Tizzzzy/Demonstration_Selection_Overview}.

\end{abstract}

%

\section{Introduction}

Large Language Models (LLMs) have achieved state-of-the-art performance across a wide range of natural language processing tasks~\cite{achiam2023gpt,dubey2024llama,AnthropicAI2023}.
One of the key factors contributing to this success is their capability for in-context learning, which allows these models to adapt to new tasks without additional training~\cite{xie2021explanation}. However, their performance is highly sensitive to the quality of the provided demonstrations. Recently, various demonstration selection algorithms have been developed to enhance this quality by selecting the most informative and relevant examples from the data pool. These algorithms have significantly reduced the time required for LLMs to address unseen tasks and have greatly improved their overall performance.

\begin{figure}
    \centering
    \includegraphics[width=1\linewidth]{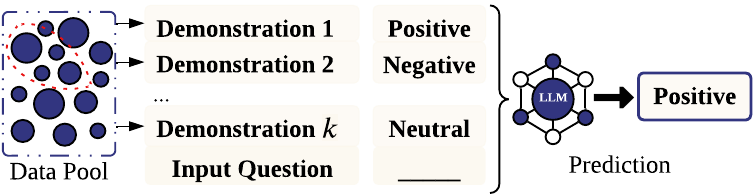}
    \caption{An overview of demonstration selection algorithms: These algorithms select demonstrations from the data pool, which the LLMs then use to generate answers.}
    \label{fig:intro}
\end{figure}

Despite the success of these approaches, the effectiveness of selected examples and the efficiency of the selection and inference processes are not well understood. This lack of understanding makes it challenging for future research to identify areas for improvement and makes it difficult to use these algorithms in real-life situations. In this paper, we present a comparative analysis of six prominent demonstration selection algorithms, evaluating them on five diverse datasets from both efficiency and effectiveness perspectives. Our study aims to provide valuable insights into the strengths and limitations of current approaches, serving as a benchmark for future research and guiding practitioners in selecting appropriate demonstration selection methods for their specific use cases.
Our comparative experiments yield the following key findings:
\begin{itemize}[leftmargin=*]\setlength\itemsep{-0.1em}
\item Contrary to expectations, not all demonstration selection algorithms consistently outperform random selection. Some algorithms struggle to surpass the random selection in certain scenarios.
\item We observe substantial accuracy gaps between different algorithms on the same dataset with the same number of demonstrations. For instance, in the MRPC dataset, this accuracy difference can be as high as 45\% when comparing CBDS with RD-direct.
\item Our analysis reveals that increasing the number of demonstrations does not always lead to better performance. The relationship between the number of demonstrations and accuracy is not monotonic and varies significantly across different tasks and algorithms.
\item We find that while some algorithms like CBDS and UPRISE achieve high accuracy, they do so at the cost of poor efficiency, requiring over 5 and 3 seconds respectively per sample for demonstration selection. This trade-off poses challenges for real-world applications where quick response times are crucial.
\end{itemize}

\section{Demonstration Selection Algorithms}
In this study, we compare six demonstration selection algorithms for in-context learning: Concept Based Demonstration Selection (CBDS), Rethinking Demonstrations direct (RD-direct) and channel (RD-channel), LLM Retriever, UPRISE, and OpenICL TopK. These algorithms employ various strategies, from leveraging latent concepts to using retrieval models, to select the best examples for LLMs. Additionally, we include OpenICL Random as a baseline, which randomly selects examples from the data pool. Here are more details of the comparing algorithms:
\begin{itemize}
    \item \textbf{C}oncept \textbf{B}ased \textbf{D}emonstration \textbf{S}election (CBDS) \cite{wang2024large}: This algorithm proposes a Bayesian approach to select demonstration examples for in-context learning in LLMs. The approach involves two main steps: (1) The optimal value of the latent concept variable \(\theta\) is learned as a set of new token embeddings using a small LLM. This step aims to align the latent concept variable with the token embedding space of the LLM. (2) After learning the optimal latent concept, the algorithm selects demonstrations that maximize the likelihood of inferring the optimal latent variable for the task at hand. These selected demonstrations can then be used with larger LLMs to improve performance. The selection process is mathematically grounded in the following equation:
    \begin{equation}
    \begin{split}
    & P^d_M(Y \mid X^d_1, Y^d_1, \ldots, X^d_k, Y^d_k, X) = \\
    & \int_\Theta P^d_M(Y \mid \theta, X) P^d_M(\theta \mid X^d_1, Y^d_1, \ldots, X^d_k, Y^d_k, X) \, d\theta.
    \end{split}
    \end{equation}
    It represents the probability of predicting the correct output \(Y\) given the selected demonstrations and the test input \(X\), integrating over the latent variable space \(\Theta\).

    \item \textbf{R}ethinking \textbf{D}emonstrations \cite{min2022rethinking}: This approach examines the factors that contribute to the success of in-context learning in LLMs. It explores the impact of different aspects of demonstrations, particularly focusing on the distribution of the input text, the label space, and the overall format. The key findings are encapsulated in the following:

    \begin{equation}
    \begin{split}
    P(y \mid x_1, y_1, \dots, x_k, y_k, x) \approx & \text{format} + \text{label space} \\
    & + \text{input distribution}
    \end{split}
    \end{equation}

    This formula reflects the idea that in-context learning performance is driven by the structure and content of the demonstrations, rather than the accuracy of the label pairings.

    \item LLM Retriever \cite{wang2023learning}: This framework focuses on selecting high-quality in-context examples to enhance the performance of LLMs. The key approach involves an iterative process where an initial set of example candidates is retrieved using an unsupervised method like BM25 \cite{robertson2009probabilistic}. These candidates are then ranked based on the conditional log probabilities of the ground truth outputs provided by the LLM. The ranking is formalized in the following equation:
    \begin{equation}
    \begin{split}
    \log p(y \mid x, x_i, y_i), \forall i \in \{1, 2, \dots, n\},
    \end{split}
    \end{equation}
    where \( p(y \mid x, x_i, y_i) \) represents the conditional probability of the output \( y \) given the input \( x \) and the \(i\)-th candidate example \((x_i, y_i)\). A reward model, based on a cross-encoder architecture, is then trained to distill these ranking preferences into a dense retriever. This retriever is further refined iteratively by leveraging the feedback from the LLM, ultimately improving the selection of in-context examples.

    \item UPRISE \cite{cheng2023uprise}: UPRISE enhances the performance of LLMs in zero-shot settings by retrieving and utilizing effective prompts. The approach involves two main steps: first, the retriever retrieves a set of positive prompts \(P^+\) from a pre-constructed pool \(P\) based on a given task input \(x\), as formulated in the equation:

    \begin{equation}
    P^+ = R(x, P),
    \end{equation}

    where \(R(x, P)\) is the retrieval function. Then, these retrieved prompts are concatenated with the input and fed into a frozen LLM to generate the output \(y\):

    \begin{equation}
    y_{P^+} = \text{LM}(y_{P^+} \mid P^+ \oplus x).
    \end{equation}

    This approach allows for cross-task and cross-model generalization, meaning the retriever is trained on diverse tasks with a smaller LLM but can be applied to larger LLMs and unseen tasks during inference.

    \item OpenICL \cite{wu2023openicl}: It is designed to facilitate ICL research and improve the evaluation of LLMs. The core approach for selecting demonstration examples involves retrieving examples based on various methods like TopK, VoteK, and BM25, which are then used to construct the context for the LLM's inference. The retrieval of examples can be formalized as:

    \begin{equation}
    (x, y) \in R(X, Y),
    \end{equation}

    where \(R(X, Y)\) represents the retrieval function applied to the training data \(X\) and \(Y\). The selected examples are then concatenated with the test input to form a single sequence, which is fed into the LLM to generate predictions. In this paper, we adopt the TopK and Random retrieval strategies.
    
\end{itemize}

\begin{figure}
    \centering
    \includegraphics[width=1\linewidth]{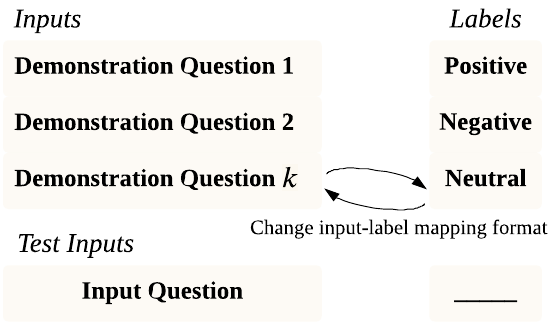}
    \caption{An visual understanding the difference between direct and channel approach}
    \label{fig:direct_channel}
\end{figure}

Notice that the RD algorithm supports both direct and channel approaches for demonstrating in-context examples, denoted as $E$. As the Figure \ref{fig:direct_channel} shows, for each example $e_i$ in $E$, it is paired with a input $x$ and a label $y$. In the direct approach, $e_i$ is structured with $x$ presented first, followed by $y$. Conversely, in the channel approach, $e_i$ is structured with $y$ presented first, followed by $x$. It is also important to note that we used ``demonstrations with gold labels'' for both the direct and channel approaches. As for other algorithms, we use the direct approach.

By evaluating these diverse approaches, we aim to provide a comprehensive analysis of current demonstration selection methods and their impact on LLM performance.

\begin{table*}[ht]
\centering
\caption{Datasets Statistics}
\label{tab:dataset}
\scalebox{0.95}{
\begin{tabular}{lccccc}
\hline
DATASETS                         & MRPC & QNLI & SST2 & CMSQA & SWAG\\ \hline
Task             & Classification & Classification & Classification & Multi-choice & Multi-choice   \\
\# Training Set         & 3,670  & 104,743 & 67,349 & 9,741 & 73,546 \\
\# Validation Set       & 408  & 5,463 & 872  & 1,221 & 20,000  \\
\# Test Set             & 1,730 & 5,463 & 1,821 & 1,284 & 20,000   \\
\hline
\end{tabular}}
\end{table*}

\section{Experiments}
In this section, we present a comprehensive evaluation of the six demonstration selection algorithms and a baseline random selection method.

\subsection{Experimental Settings}
\subsubsection{Datasets:} The demonstration selection algorithms will be evaluated on 5 datasets, categorized into two groups: classification and multi-choice. The classification datasets include GLUE-MRPC, GLUE-QNLI, and GLUE-SST2. The multi-choice datasets are CMSQA and SWAG. Their statistic is shown in Table \ref{tab:dataset}, and below are more details of the datasets:
\begin{itemize}
    \item GLUE-MRPC \cite{wang2018glue}: MRPC is a dataset consisting of sentence pairs, each annotated with a binary label indicating whether the sentences in the pair are semantically equivalent.
    
    \item GLUE-QNLI: QNLI dataset is derived from the Stanford Question Answering Dataset (SQuAD). It is a binary classification task where the goal is to determine whether the context sentence contains the answer to the question.

    \item GLUE-SST2: SST-2 is a binary sentiment classification dataset that includes movie review excerpts, where the task is to predict whether the sentiment of the review is positive or negative.

    \item CMSQA \cite{talmor2018commonsenseqa}: The CommonsenseQA dataset consists of multiple-choice questions that require commonsense knowledge. Each question is paired with five answer choices, where only one is correct.

    \item SWAG \cite{zellers2018swag}: SWAG dataset is a large-scale benchmark for grounded commonsense inference. It contains multiple-choice questions about grounded situations, where the task is to select the most plausible continuation of a given scenario.
\end{itemize}

\begin{figure}
    \centering
    \includegraphics[width=1\linewidth]{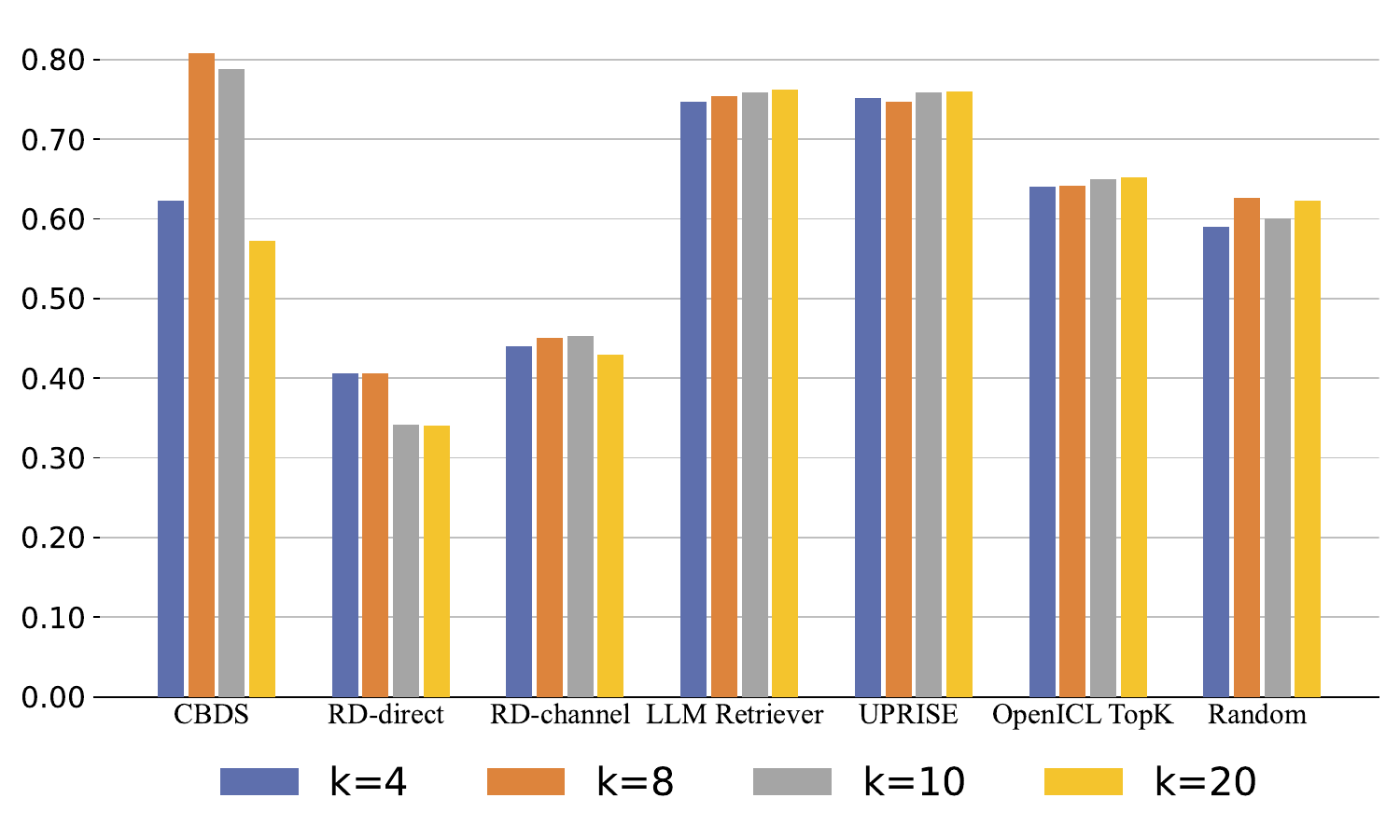}
    \caption{The effectiveness of the algorithms on MRPC dataset}
    \label{fig:effectiveness}
\end{figure}

\subsubsection{Metrics:} Given that all datasets are either classification or multi-choice, accuracy is an appropriate metric for evaluating the performance. To show the algorithm's computational efficiency, we present the results in seconds.

\subsubsection{Implementation Details:} 

All of the demonstration selection algorithms were tested on LLaMa3-8B \cite{touvron2023llama} to evaluate their effectiveness. As shown in Figure \ref{fig:intro}, each algorithm first selects $k$ demonstration examples from the available data pool based on the input question, which serve as in-context learning exemplars. These exemplars, along with the input question, are then fed into LLaMa3 for evaluation. We measured the absolute time each algorithm took to select $k$ demonstration examples per single input, as well as the model inference time for the same input. To ensure fairness, we excluded all offline computational time, such as pre-training or other preliminary computations, for all algorithms. The values of $k$ considered in this study were \{4, 8, 10, 20\}. All experiments were conducted using an NVIDIA RTX A6000 GPU, with the LLaMa3-8B model directly loaded from Huggingface, and we used the default sampling parameters such as temperature and top\_p.

\begin{figure*}
    \centering
    \includegraphics[width=0.8\linewidth]{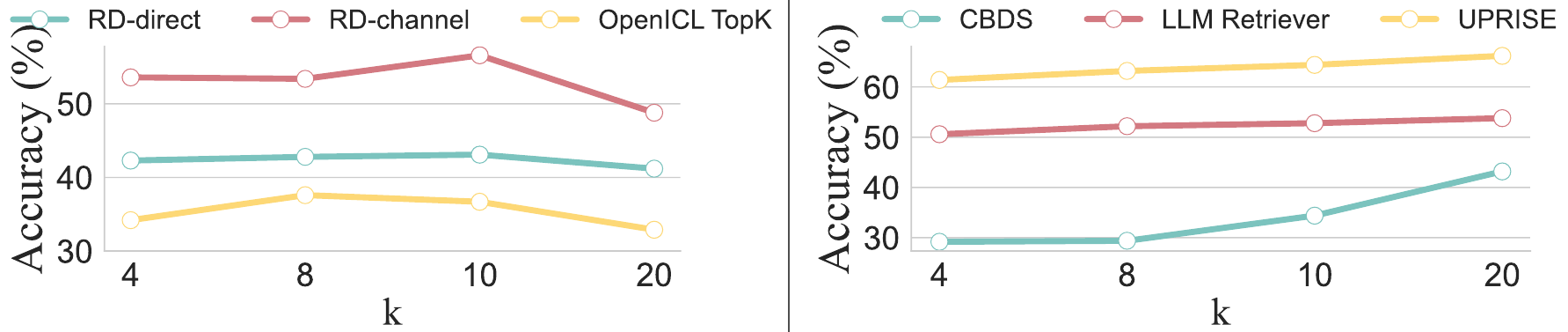}
    \caption{An visualize trend of the algorithms effectiveness}
    \label{fig:trend}
\end{figure*}

\begin{table*}[ht]
\centering
\caption{Effectiveness of the Demonstration Selection Algorithms}
\setlength{\tabcolsep}{3pt}  
\scalebox{0.85}{
\begin{tabular}{l*{5}{cccc}}
\toprule
{Datasets} & \multicolumn{4}{c}{MRPC} & \multicolumn{4}{c}{QNLI} & \multicolumn{4}{c}{CMSQA} & \multicolumn{4}{c}{SWAG} & \multicolumn{4}{c}{SST2} \\
\cmidrule(lr){2-5} \cmidrule(lr){6-9} \cmidrule(lr){10-13} \cmidrule(lr){14-17} \cmidrule(lr){18-21}
 & k=4 & k=8 & k=10 & k=20 & k=4 & k=8 & k=10 & k=20 & k=4 & k=8 & k=10 & k=20 & k=4 & k=8 & k=10 & k=20 & k=4 & k=8 & k=10 & k=20 \\
\midrule
CBDS            & 62.3 & \textbf{80.8} & \textbf{78.8} & 57.3 & 51.1 & 50.7 & 54.4 & 54.8 & 29.2 & 29.4 & 34.4 & 43.2 & 36.8 & 38.1 & 34.6 & 36.9 & 91.4 & \textbf{93.9} & 93.5 & 73.2 \\
RD-direct       & 40.6 & 40.6 & 34.2 & 34.0 & 33.6 & 35.1 & 35.9 & 35.5 & 42.3 & 42.8 & 43.1 & 41.2 & 56.2 & 56.6 & 57.3 & 56.4 & 91.9 & 75.7 & \textbf{94.6} & 83.8 \\
RD-channel      & 44.0 & 45.1 & 45.3 & 42.9 & 46.7 & 47.8 & 44.9 & 38.8 & 53.6 & 53.4 & 56.6 & 48.8 & 53.2 & 52.0 & 51.8 & 57.8 & 91.2 & 87.9 & 79.3 & 76.4 \\
LLM Retriever   & 74.7 & 75.4 & 75.9 & \textbf{76.2} & 71.5 & 72.3 & 72.9 & 72.7 & 50.6 & 52.2 & 52.8 & 53.8 & 65.1 & 64.7 & 64.9 & \textbf{64.7} & \textbf{93.7} & 93.8 & 93.6 & 93.5 \\
UPRISE          & \textbf{75.2} & 74.7 & 75.9 & 76.0 & \textbf{81.8} & \textbf{81.9} & \textbf{81.8} & \textbf{81.1} & \textbf{61.4} & \textbf{63.2} & \textbf{64.4} & \textbf{66.2} & \textbf{69.0} & \textbf{66.8} & \textbf{69.9} & 61.9 & 93.4 & 92.5 & 92.4 & 91.3 \\
OpenICL TopK    & 64.1 & 64.2 & 65.0 & 65.2 & 78.1 & 74.7 & 70.8 & 69.9 & 34.2 & 37.6 & 36.7 & 32.9 & 35.4 & 36.7 & 37.1 & 34.9 & 91.2 & 92.1 & 92.3 & \textbf{94.5} \\
\midrule
Random Selection  & 59.0 & 62.6 & 60.1 & 62.3 & 51.8 & 48.3 & 47.9 & 47.4 & 26.7 & 28.1 & 28.1 & 28.5 & 29.6 & 31.8 & 31.4 & 31.1 & 86.2 & 89.5 & 91.3 & 91.7 \\
\bottomrule
\end{tabular}}
\label{table:effectiveness}
\end{table*}

\subsection{Main Results}
Table \ref{table:effectiveness} presents the effectiveness of each demonstration selection algorithm across different datasets and varying numbers of $k$ in-context examples. The highest accuracy in each column is highlighted in bold format. We evaluated the computational efficiency of the six algorithms and a baseline, and report the results in Table \ref{table:efficiency}. We will show our findings from two perspectives: effectiveness and efficiency.

\subsubsection{Effectiveness Perspective: }
Contrary to expectations, not all demonstration selection algorithms consistently outperform random selection. While some algorithms show significant improvements, others struggle to surpass the baseline set by random selection in certain scenarios. For instance, as shown in Table \ref{table:effectiveness}, we observed that for the MRPC dataset, both the RD-direct and RD-channel algorithms demonstrate limited effectiveness compared to random selection. This finding challenges the assumption that sophisticated selection methods always yield better results and highlights the importance of thorough evaluation before implementation. Another notable observation arises when we plot the effectiveness of each algorithm on the MRPC dataset, as shown in Figure \ref{fig:effectiveness}. There is a substantial accuracy gap between different algorithms on the same dataset with the same $k$ value. In the MRPC dataset, this accuracy difference can be as high as 45\%, as seen when comparing CBDS with $k=10$ to RD-direct with $k=10$.

A closer examination of the SST2 dataset in Table \ref{table:effectiveness} reveals that for simpler classification tasks, the effectiveness of demonstration selection algorithms is often limited and does not significantly improve accuracy compared to more challenging multi-choice tasks like CommonsenseQA and SWAG.

Furthermore, the RD algorithm provides both direct and channel approaches for presenting demonstration examples to the model. In our comparisons, we used ``demonstrations with gold labels'' for both the direct and channel approaches. We found that the channel approach generally outperforms the direct approach. This observation aligns with the results presented in Figures 8-10 of the original paper \cite{min2022rethinking}. A possible explanation for this is that the channel approach better aligns the input features with the class labels, making it easier for the model to capture and understand the underlying relationships.

Our analysis also reveals that increasing the number of demonstrations ($k$) does not always lead to better performance. The relationship between $k$ and accuracy is not monotonic and varies significantly across tasks and algorithms. In Figure \ref{fig:trend}, we visualize the effectiveness trends of six algorithms on the CMSQA dataset. We observed two primary patterns: (1) as shown on the left, performance peaks at a certain $k$ before declining, and (2) on the right, accuracy continues to improve as $k$ increases. For the pattern on the left, we hypothesize that for some tasks, an excessive number of in-context examples may introduce conflicting knowledge, either with the model's internal knowledge or among the examples themselves. Additionally, some examples might be irrelevant to the input question and may act as noise. This finding underscores the importance of selecting an appropriate $k$ value tailored to the specific task. Thus, it is crucial to tune the number of demonstrations for each specific task and algorithm combination.

\begin{table*}[ht]
\centering
\caption{Efficiency of the Demonstration Selection Algorithms}
\scalebox{0.80}{
\begin{tabular}{l*{5}{cc}}
\toprule
{Datasets} & \multicolumn{2}{c}{MRPC} & \multicolumn{2}{c}{QNLI} & \multicolumn{2}{c}{CMSQA} & \multicolumn{2}{c}{SWAG} & \multicolumn{2}{c}{SST2} \\
\cmidrule(lr){2-3} \cmidrule(lr){4-5} \cmidrule(lr){6-7} \cmidrule(lr){8-9} \cmidrule(lr){10-11}
 & selection & inference & selection & inference & selection & inference & selection & inference & selection & inference\\
\midrule
CBDS            & 5.44   & 1.82   & 5.35   & 1.80  & 5.43 & 4.58   & 5.45   & 3.63   & 5.44  & 1.82 \\
RD-direct       & $2.34*10^{-3}$   & 1.80   & $2.37*10^{-3}$   & 1.81  & $2.44*10^{-3}$ & 4.56   & $2.50*10^{-3}$   & 3.63   & $1.44*10^{-3}$  & 1.78\\
RD-channel      & $2.24*10^{-3}$   & 1.80   & $2.50*10^{-3}$   & 1.79  & $2.49*10^{-3}$ & 4.55   & $2.46*10^{-3}$   & 3.62   & $1.44*10^{-3}$  & 1.77\\
LLM Retriever   & $8.32*10^{-1}$   & 12.81   & $6.74*10^{-1}$   & 12.84  & $8.47*10^{-1}$ & 12.87   & $8.83*10^{-1}$   & 11.85   & $8.16*10^{-1}$  & 12.62\\
UPRISE          & 3.06   & 3.67   & 3.23   & 3.66  & 3.25 & 3.69   & 3.09   & 3.68   & 3.62  & 3.43\\
OpenICL TopK    & $1.46*10^{-1}$   & 1.20   & $2.22*10^{-1}$   & 1.19  & $1.48*10^{-1}$ & 2.94   & $2.60*10^{-1}$   & 2.61   & $4.79*10^{-1}$  & 0.81\\
\midrule
OpenICL Random  & $5.24*10^{-6}$   & 1.11   & $5.63*10^{-6}$   & 1.10  & $5.34*10^{-6}$ & 3.04   & $6.48*10^{-6}$   & 2.61   & $8.11*10^{-6}$  & 0.61\\
\bottomrule
\label{table:efficiency}
\end{tabular}}
\end{table*}

\subsubsection{Efficiency Perspective: }
We also evaluated the computational efficiency of the six algorithms and a baseline. As shown in Table \ref{table:efficiency}, we measured efficiency in two ways: (1) the absolute demonstration selection time (selection) required for an algorithm to select $k$ in-context examples for a single input question, and (2) the absolute inference time (inference) required for the model to generate an answer for the same input question. All efficiency experiments were conducted with $k=10$.

Among the compared algorithms, CBDS and UPRISE stand out for their high accuracy but also for their poor efficiency. On average, these algorithms require over 5 seconds and 3 seconds respectively per sample for demonstration selection. Referring to Table \ref{table:effectiveness}, out of 20 combinations (4 $k$ values × 5 datasets), CBDS or UPRISE achieved the best accuracy in 14 cases. However, this superior accuracy comes at the cost of extremely high computational complexity. The low efficiency of these algorithms makes them challenging to apply in real-world applications where quick response times are crucial.

We also found out that five algorithms are relatively efficient, taking less than one second per input question to select 10 examples. Notably, the OpenICL Random algorithm, along with both the RD-direct and RD-channel approaches, demonstrated particularly fast demonstration selection times. However, when cross-referencing these results with Table \ref{table:effectiveness}, it becomes clear that this speed often comes at a cost to accuracy. These faster algorithms generally do not perform as effectively as others. On the other hand, the LLM Retriever achieves a more balanced trade-off, offering reasonable demonstration selection speed while maintaining good overall effectiveness.

We also observed variations in inference times across different algorithms. As shown in the `inference' column, the CBDS and RD algorithms exhibit similar inference times, as both utilize model inference code from the MetaICL codebase \cite{min2021metaicl}. An interesting finding emerges when comparing inference times for different datasets within the same algorithm: for most algorithms, including CBDS, RD, and OpenICL, inference times increase for more complex tasks, such as CMSQA and SWAG, compared to simpler binary classification tasks. 

As for the LLM Retriever, although it demonstrates fast demonstration selection times, the overall computational time remains slower when factoring in inference time. This slowdown is likely due to the constrained generation technique employed by the algorithm \cite{celikyilmaz2020evaluation}. Specifically, when employing constrained generation with a prefix trie, the model needs to check each generated token against the trie to ensure it aligns with the allowed prefixes. This additional step can slow down the generation process because it introduces extra computational overhead to enforce these constraints. This finding suggests that future research should not only focus on improving demonstration selection efficiency but also consider optimizing model inference time to achieve better overall computational performance.

\section{Related Work}
In-context learning has emerged as a powerful capability of LLMs, enabling them to adapt to new tasks without fine-tuning, but the effectiveness of this approach heavily depends on the quality and relevance of the provided demonstrations.
With numerous demonstration selection algorithms in the field, the optimal method for selecting these examples remains an open question. For instance, \citet{wang2024large} emphasize aligning demonstrations with the model's latent structure, improving performance by uncovering latent variable explanations. \citet{zhang2022active, qin2023context} employ an active and iterative selection approach to identify the most informative examples. Retrieval-based methods are utilized by \citet{wang2023learning, wang2024mdr, li2023unified, xu2024context} to select demonstrations. \citet{liu2024demorank} adopt a ranking mechanism that considers relevance, similarity, and diversity for optimal demonstration selection. \citet{ye2023compositional} suggest constructing demonstrations through the combination of simpler examples to better capture complex task dynamics. \citet{van2024context} introduce InfICL, which uses influence functions to identify the most impactful examples for selection. \citet{liu2024unraveling} uncover the importance of task-agnostic multi-level similarity and task-specific label similarities, proposing methods that integrate these factors for improved performance across diverse tasks. Additionally, \citet{kim2022self, zhou2024enhancing} introduce the concept of self-generated demonstrations, which involves generating examples. Together, these works underscore the critical role of demonstration selection in enhancing the effectiveness of in-context learning across varied applications.

Our work differs from previous studies by providing a comparative analysis of multiple demonstration selection algorithms across various tasks and datasets. While existing literature has primarily focused on developing individual algorithms or comparing a limited set of methods, we offer a holistic evaluation of both the effectiveness and efficiency of six demonstration selection algorithms.

\section{Conclusions and Future Work}

In this paper, our comprehensive evaluation of demonstration selection algorithms reveals significant variations in both effectiveness and efficiency across different tasks and datasets. While some algorithms show promising results, the trade-offs between accuracy and computational cost present challenges for real-world applications. Our findings highlight the need for task-specific optimization and suggest potential avenues for future research. These include developing adaptive algorithms that can dynamically adjust the number of demonstrations based on task complexity, integrating more advanced retrieval techniques, and exploring methods to balance effectiveness with computational efficiency. 
\section*{Acknowledgment}
The work is in part supported by NSF \#2310261. The views and conclusions in this paper are those of the authors and should not be interpreted as representing funding agencies.

\bibliography{aaai25}

\begin{thebibliography}{27}
\providecommand{\natexlab}[1]{#1}

\bibitem[{Achiam et~al.(2023)Achiam, Adler, Agarwal, Ahmad, Akkaya, Aleman, Almeida, Altenschmidt, Altman, Anadkat et~al.}]{achiam2023gpt}
Achiam, J.; Adler, S.; Agarwal, S.; Ahmad, L.; Akkaya, I.; Aleman, F.~L.; Almeida, D.; Altenschmidt, J.; Altman, S.; Anadkat, S.; et~al. 2023.
\newblock Gpt-4 technical report.
\newblock \emph{arXiv preprint arXiv:2303.08774}.

\bibitem[{AnthropicAI(2023)}]{AnthropicAI2023}
AnthropicAI. 2023.
\newblock Introducing claude.

\bibitem[{Celikyilmaz, Clark, and Gao(2020)}]{celikyilmaz2020evaluation}
Celikyilmaz, A.; Clark, E.; and Gao, J. 2020.
\newblock Evaluation of text generation: A survey.
\newblock \emph{arXiv preprint arXiv:2006.14799}.

\bibitem[{Cheng et~al.(2023)Cheng, Huang, Bi, Zhan, Liu, Wang, Sun, Wei, Deng, and Zhang}]{cheng2023uprise}
Cheng, D.; Huang, S.; Bi, J.; Zhan, Y.; Liu, J.; Wang, Y.; Sun, H.; Wei, F.; Deng, D.; and Zhang, Q. 2023.
\newblock Uprise: Universal prompt retrieval for improving zero-shot evaluation.
\newblock \emph{arXiv preprint arXiv:2303.08518}.

\bibitem[{Dubey et~al.(2024)Dubey, Jauhri, Pandey, Kadian, Al-Dahle, Letman, Mathur, Schelten, Yang, Fan et~al.}]{dubey2024llama}
Dubey, A.; Jauhri, A.; Pandey, A.; Kadian, A.; Al-Dahle, A.; Letman, A.; Mathur, A.; Schelten, A.; Yang, A.; Fan, A.; et~al. 2024.
\newblock The llama 3 herd of models.
\newblock \emph{arXiv preprint arXiv:2407.21783}.

\bibitem[{Kim et~al.(2022)Kim, Cho, Kim, Kim, Yoo, and Lee}]{kim2022self}
Kim, H.~J.; Cho, H.; Kim, J.; Kim, T.; Yoo, K.~M.; and Lee, S.-g. 2022.
\newblock Self-generated in-context learning: Leveraging auto-regressive language models as a demonstration generator.
\newblock \emph{arXiv preprint arXiv:2206.08082}.

\bibitem[{Li et~al.(2023)Li, Lv, Yan, Lin, Zhu, Ni, Xie, Wang, and Qiu}]{li2023unified}
Li, X.; Lv, K.; Yan, H.; Lin, T.; Zhu, W.; Ni, Y.; Xie, G.; Wang, X.; and Qiu, X. 2023.
\newblock Unified demonstration retriever for in-context learning.
\newblock \emph{arXiv preprint arXiv:2305.04320}.

\bibitem[{Liu et~al.(2024)Liu, Wang, Sun, Tian, Kong, Dong, and Li}]{liu2024unraveling}
Liu, H.; Wang, W.; Sun, H.; Tian, C.~X.; Kong, C.; Dong, X.; and Li, H. 2024.
\newblock Unraveling the Mechanics of Learning-Based Demonstration Selection for In-Context Learning.
\newblock \emph{arXiv preprint arXiv:2406.11890}.

\bibitem[{Liu, Zhu, and Dou(2024)}]{liu2024demorank}
Liu, W.; Zhu, Y.; and Dou, Z. 2024.
\newblock DemoRank: Selecting Effective Demonstrations for Large Language Models in Ranking Task.
\newblock \emph{arXiv preprint arXiv:2406.16332}.

\bibitem[{Min et~al.(2021)Min, Lewis, Zettlemoyer, and Hajishirzi}]{min2021metaicl}
Min, S.; Lewis, M.; Zettlemoyer, L.; and Hajishirzi, H. 2021.
\newblock Metaicl: Learning to learn in context.
\newblock \emph{arXiv preprint arXiv:2110.15943}.

\bibitem[{Min et~al.(2022)Min, Lyu, Holtzman, Artetxe, Lewis, Hajishirzi, and Zettlemoyer}]{min2022rethinking}
Min, S.; Lyu, X.; Holtzman, A.; Artetxe, M.; Lewis, M.; Hajishirzi, H.; and Zettlemoyer, L. 2022.
\newblock Rethinking the role of demonstrations: What makes in-context learning work?
\newblock \emph{arXiv preprint arXiv:2202.12837}.

\bibitem[{Qin et~al.(2023)Qin, Zhang, Dagar, and Ye}]{qin2023context}
Qin, C.; Zhang, A.; Dagar, A.; and Ye, W. 2023.
\newblock In-context learning with iterative demonstration selection.
\newblock \emph{arXiv preprint arXiv:2310.09881}.

\bibitem[{Robertson, Zaragoza et~al.(2009)}]{robertson2009probabilistic}
Robertson, S.; Zaragoza, H.; et~al. 2009.
\newblock The probabilistic relevance framework: BM25 and beyond.
\newblock \emph{Foundations and Trends{\textregistered} in Information Retrieval}, 3(4): 333--389.

\bibitem[{Talmor et~al.(2018)Talmor, Herzig, Lourie, and Berant}]{talmor2018commonsenseqa}
Talmor, A.; Herzig, J.; Lourie, N.; and Berant, J. 2018.
\newblock Commonsenseqa: A question answering challenge targeting commonsense knowledge.
\newblock \emph{arXiv preprint arXiv:1811.00937}.

\bibitem[{Touvron et~al.(2023)Touvron, Martin, Stone, Albert, Almahairi, Babaei, Bashlykov, Batra, Bhargava, Bhosale et~al.}]{touvron2023llama}
Touvron, H.; Martin, L.; Stone, K.; Albert, P.; Almahairi, A.; Babaei, Y.; Bashlykov, N.; Batra, S.; Bhargava, P.; Bhosale, S.; et~al. 2023.
\newblock Llama 2: Open foundation and fine-tuned chat models.
\newblock \emph{arXiv preprint arXiv:2307.09288}.

\bibitem[{Van, Wu et~al.(2024)}]{van2024context}
Van, M.-H.; Wu, X.; et~al. 2024.
\newblock In-Context Learning Demonstration Selection via Influence Analysis.
\newblock \emph{arXiv preprint arXiv:2402.11750}.

\bibitem[{Wang et~al.(2018)Wang, Singh, Michael, Hill, Levy, and Bowman}]{wang2018glue}
Wang, A.; Singh, A.; Michael, J.; Hill, F.; Levy, O.; and Bowman, S.~R. 2018.
\newblock GLUE: A multi-task benchmark and analysis platform for natural language understanding.
\newblock \emph{arXiv preprint arXiv:1804.07461}.

\bibitem[{Wang et~al.(2024{\natexlab{a}})Wang, Wu, Sun, Xia, Cheng, Wang, Qi, and Liao}]{wang2024mdr}
Wang, H.; Wu, J.; Sun, H.; Xia, Z.; Cheng, D.; Wang, J.; Qi, Q.; and Liao, J. 2024{\natexlab{a}}.
\newblock MDR: Model-Specific Demonstration Retrieval at Inference Time for In-Context Learning.
\newblock In \emph{Proceedings of the 2024 Conference of the North American Chapter of the Association for Computational Linguistics: Human Language Technologies (Volume 1: Long Papers)}, 4189--4204.

\bibitem[{Wang, Yang, and Wei(2023)}]{wang2023learning}
Wang, L.; Yang, N.; and Wei, F. 2023.
\newblock Learning to retrieve in-context examples for large language models.
\newblock \emph{arXiv preprint arXiv:2307.07164}.

\bibitem[{Wang et~al.(2024{\natexlab{b}})Wang, Zhu, Saxon, Steyvers, and Wang}]{wang2024large}
Wang, X.; Zhu, W.; Saxon, M.; Steyvers, M.; and Wang, W.~Y. 2024{\natexlab{b}}.
\newblock Large language models are latent variable models: Explaining and finding good demonstrations for in-context learning.
\newblock \emph{Advances in Neural Information Processing Systems}, 36.

\bibitem[{Wu et~al.(2023)Wu, Wang, Ye, Feng, Xu, Qiao, and Wu}]{wu2023openicl}
Wu, Z.; Wang, Y.; Ye, J.; Feng, J.; Xu, J.; Qiao, Y.; and Wu, Z. 2023.
\newblock Openicl: An open-source framework for in-context learning.
\newblock \emph{arXiv preprint arXiv:2303.02913}.

\bibitem[{Xie et~al.(2021)Xie, Raghunathan, Liang, and Ma}]{xie2021explanation}
Xie, S.~M.; Raghunathan, A.; Liang, P.; and Ma, T. 2021.
\newblock An explanation of in-context learning as implicit bayesian inference.
\newblock \emph{arXiv preprint arXiv:2111.02080}.

\bibitem[{Xu et~al.(2024)Xu, Liu, Pasupat, Kazemi et~al.}]{xu2024context}
Xu, X.; Liu, Y.; Pasupat, P.; Kazemi, M.; et~al. 2024.
\newblock In-context learning with retrieved demonstrations for language models: A survey.
\newblock \emph{arXiv preprint arXiv:2401.11624}.

\bibitem[{Ye et~al.(2023)Ye, Wu, Feng, Yu, and Kong}]{ye2023compositional}
Ye, J.; Wu, Z.; Feng, J.; Yu, T.; and Kong, L. 2023.
\newblock Compositional exemplars for in-context learning.
\newblock In \emph{International Conference on Machine Learning}, 39818--39833. PMLR.

\bibitem[{Zellers et~al.(2018)Zellers, Bisk, Schwartz, and Choi}]{zellers2018swag}
Zellers, R.; Bisk, Y.; Schwartz, R.; and Choi, Y. 2018.
\newblock Swag: A large-scale adversarial dataset for grounded commonsense inference.
\newblock \emph{arXiv preprint arXiv:1808.05326}.

\bibitem[{Zhang, Feng, and Tan(2022)}]{zhang2022active}
Zhang, Y.; Feng, S.; and Tan, C. 2022.
\newblock Active example selection for in-context learning.
\newblock \emph{arXiv preprint arXiv:2211.04486}.

\bibitem[{Zhou et~al.(2024)Zhou, Ye, Wang, Jiang, Lee, Xie, and Zhang}]{zhou2024enhancing}
Zhou, X.; Ye, W.; Wang, Y.; Jiang, C.; Lee, Z.; Xie, R.; and Zhang, S. 2024.
\newblock Enhancing In-Context Learning via Implicit Demonstration Augmentation.
\newblock \emph{arXiv preprint arXiv:2407.00100}.

\end{thebibliography}

\end{document}